\documentclass[sigconf]{acmart}

\usepackage{amsfonts}
\usepackage{amsmath}
\usepackage{booktabs}
\usepackage{multirow}
\usepackage{algorithm}
\usepackage{algorithmic}
\usepackage{subfigure}
\usepackage[bottom]{footmisc}

\AtBeginDocument{%
  \providecommand\BibTeX{{%
    \normalfont B\kern-0.5em{\scshape i\kern-0.25em b}\kern-0.8em\TeX}}}

\setcopyright{acmcopyright}
\copyrightyear{XXXX}
\acmYear{XXXX}
\acmDOI{XXXX/XXXX.XXXX}

\acmConference[XXXX]{XXXX: ACM Symposium on Neural
  Gaze Detection}{XXXX, XXXX}{XXXX}
\acmBooktitle{XXXX XXXX: XXXX,
  XXXX}
\acmPrice{15.00}
\acmISBN{978-1-4503-XXXX-X/21/04}

\author{Ning Ma$^{1}$, Jiajun Bu$^{1}$, Lixian Lu$^{1}$, Jun Wen $^{2}$, Zhen Zhang$^{1}$}
\author{Sheng Zhou$^{1}$, Xifeng Yan$^{3}$}
\affiliation{%
  \institution{
  \textsuperscript{\rm 1} College of Computer Science, Zhejiang University \\
  \textsuperscript{\rm 2} Department of Biomedical Informatics, Harvard Medical School\\
  \textsuperscript{\rm 3} Computer Science Department, University of California Santa Barbara
  }
}
\email{{ma\_ning, bjj,llx97, junwen ,zhousheng\_zju, zhen\_zhang}@zju.edu.cn, xyan@cs.ucsb.edu}

\begin{document}

\title{Semi-Supervised Hypothesis Transfer for Source-Free Domain Adaptation}

\renewcommand{\shortauthors}{Trovato and Tobin, et al.}

\begin{abstract}
Domain Adaptation has been widely used to deal with the distribution shift in vision, language, multimedia etc. Most domain adaptation methods learn domain-invariant features with data from both domains available. However, such a strategy might be infeasible in practice when source data are unavailable due to data-privacy concerns. To address this issue, we propose a novel adaptation method via hypothesis transfer without accessing source  data at adaptation stage. In order to fully use the limited target data, a semi-supervised mutual enhancement method is proposed, in which entropy minimization and augmented label propagation are used iteratively to perform inter-domain and intra-domain alignments. Compared with state-of-the-art methods, the experimental results on three public datasets demonstrate that our method gets up to 19.9\% improvements on semi-supervised adaptation tasks. 
\end{abstract}




\maketitle

\section{Introduction}
Recognizing and classifying images is a fundamental visual task in machine learning community.  
Although Deep Neural Networks (DNNs)  have achieved excellent performances on this task, it suffers significant degradation when there is \textit{distribution shift} in image data. 
To overcome this issue, Domain Adaptation (DA) \cite{7078994,UDA_Survey} is proposed by generalizing knowledge learned from the source domain to the target domain, with success achieved in varied areas, \emph{eg.}, computer vision \cite{ganin2015unsupervised,MME} and multi-media \cite{6971165,806234,3123429}. 
Most existing DA methods assume that the source domain data are labeled while the target domain data are unlabeled.
However, a few labeled target data may be additionally available in practice and beneficial for the cross-domain knowledge transfer.
Learning on the target domain with knowledge from both labeled source data and partially labeled target data is also known as the Semi-Supervised Domain Adaptation (SSDA) \cite{NIPS20104009,MME}.

\begin{figure}[!tp]
\centering
\includegraphics[width=1\columnwidth]{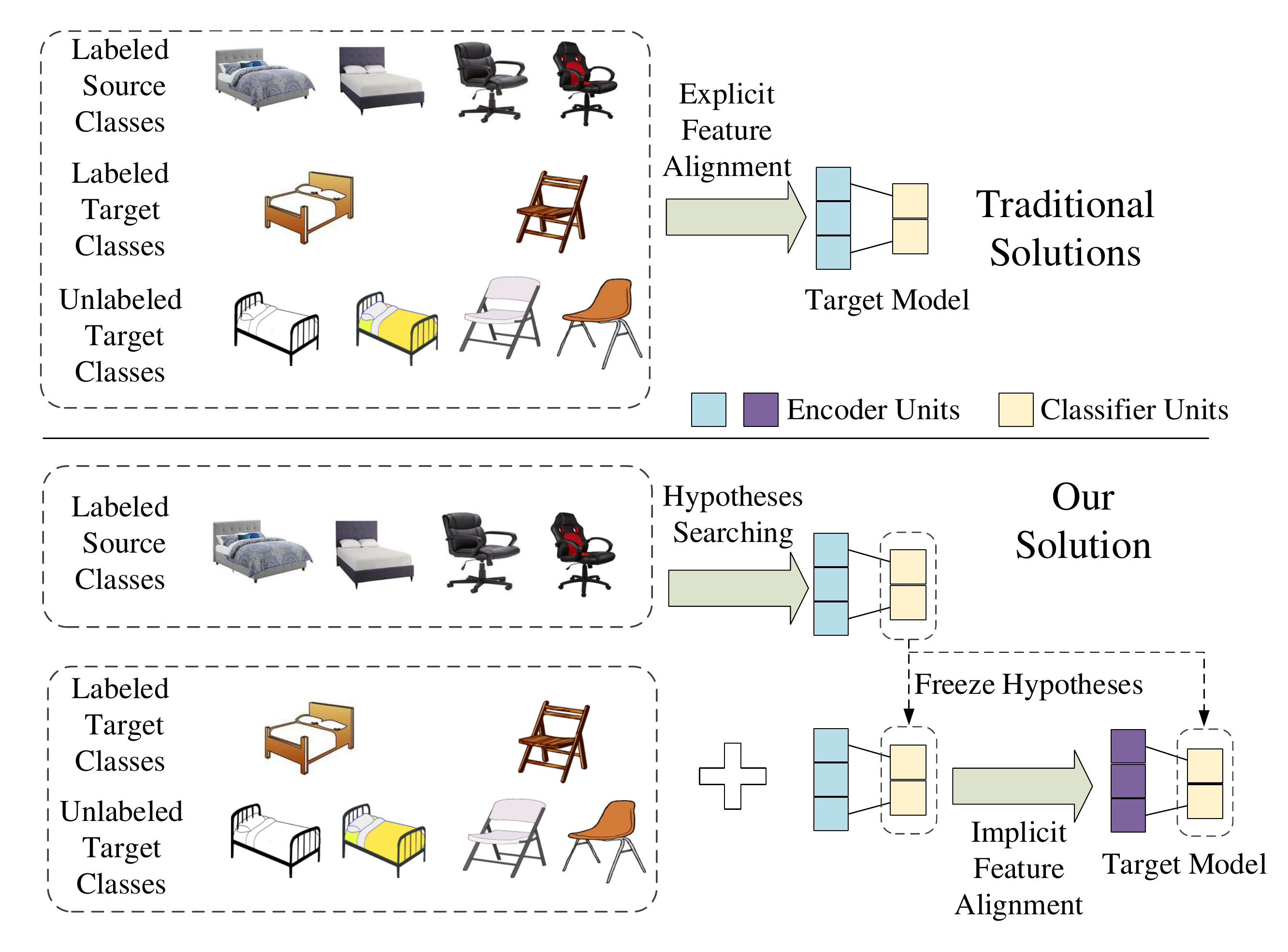} 
\caption{Comparison between traditional SSDA approaches and the proposed method. The source data are unseen at adaptation stage. On the target domain, only the encoder modules are updated for implicit feature alignment.}
\label{fig:motivation}
\end{figure}

Due to the practical significance, SSDA is receiving increasing attention recently \cite{MME,BiAT,kim2020attract,li2020online,qin2020opposite}. Among these SSDA methods, there are mainly two paradigms according to different feature alignment considerations. 1) Inter-domain alignment \cite{MME}, in which feature alignment is merely performed between the unlabeled target data and labeled source data while the labeled target data is used in a supervised manner. 2) Intra-domain alignment, the feature discrepancy between the labeled and unlabeled target data is further considered, which is shown to reduce domain shift considerably \cite{BiAT,kim2020attract}. Despite of the success, these methods assume simultaneous availability of source and target data, which could be infeasible in practice when source data is unavailable due to data privacy policy. Take the medical data as an example, the source data and target data may be possessed by different hospitals which are inaccessible from each other.

Although important, performing adaptation on the partially labeled  target data without accessing the source data is of great challenge to the SSDA task.  
Specifically, based on the setting of inaccessible source data at the adaptation stage,  \textit{how to align the target representations with the source domain?  how to make full use of a few labeled target data?} 
Recently, several works have been proposed for source-free domain adaptation without accessing the source data (Tent \cite{wang2021tent}, SHOT \cite{SHOT}, MAda\cite{Li_2020_CVPR}), but few of them construct specialized method for the semi-supervised setting. 
For the inter-domain alignment, 
Tent \cite{wang2021tent} and SHOT \cite{SHOT} propose implicit feature alignment methods by updating the normalization layers or encoder layers, while lacking the enhancement mechanism using labeled target data. Furthermore, without augmentation, one or three labeled target data  provide limited effect to enhance the inter-domain and intra-domain alignments.

To tackle the above challenges, we introduce our SSDA solution in the source-free setting. Different from the traditional SSDA methods, we train a model on the source domain and only transfer the model to the target domain to satisfy the privacy policy. 
The model is assumed to have captured appropriate class hypotheses from the source domain, which is critical for the target domain. 
Hence at the adaptation stage, a \textit{frozen} classifier module with an \textit{adjustable} feature encoder derived from source data is used via the framework of Hypothesis Transfer Learning (HTL \cite{HTL}).
HTL has been explored empirically and theoretically in transfer learning \cite{feifei, HTL, SHOT}, which aims to find a proper hypothesis (classifier module) derived from the source domain and then apply it to the target domain.   Figure \ref{fig:motivation} illustrates the differences between traditional SSDA solutions and our method. 

Concretely, we firstly consider implicit feature alignment on the partially labeled target domain when given the classification hypothesis derived from source domain.  To approach the representations of source data, the representations of target data should stay away from the frozen classification boundary. 
We implement this setting using Entropy Minimization \cite{EM} on the target domain. 
However, this will enforce the model to be over confident with the misclassified data. 
To alleviate the problem, we further propose the augmented Label Propagation  (LP \cite{LP}) using uncertainty estimation \cite{rizve2021in,blundell15}. 
The augmented LP enforces a few ground truth labels and high quality pseudo-labels of target data to span the fixed classification boundary and propagate label information to the unlabeled target data.  
Moreover, when Entropy Minimization is applied to the unlabeled target data,  the data with lower entropy are usually allocated with high quality pseudo labels. These labels further augment the labeled target data and enhance the label propagation to improve the intra-domain alignment.  
Besides, the representations of low entropy target data usually have better alignment with the source representations, which also improves the cross-domain alignment. 
By building the mutual  enhancement between the entropy minimization and label propagation, the empirical performances for SSDA get promise improvements.   

 The experiments on three datasets demonstrate that our method gets state-of-the-art performance compared to  SSDA baselines.  Our method can be formulated as \textbf{M}utual \textbf{E}nhancement training for \textbf{S}emi-supervised \textbf{H}ypothesis transfer (\textbf{MESH}), and the contributions are summarized as follows:
\begin{itemize}
\item Existing SSDA methods usually need to access the source domain data when performing adaptation on the target domain, which sacrifices the privacy preserving for source domain data. We propose a privacy preserving method for the SSDA problem via semi-supervised hypothesis transfer. 
\item To alleviate the limitation of  entropy minimization, a pseudo labeling loss driven by label propagation is used to corrects the gradient direction of  entropy minimization loss. Further, to augment the labeled target data, a few low entropy data which have experienced entropy minimization are utilized to boost the  label propagation process. With such a mutual enhancement,  the entropy minimization and label propagation can promote each other.
\item The empirical experiments show that our method gets state-of-the-art performances with  up to 19.9\% improvement over recent SSDA baselines, while accomplishing well privacy preserving for source data.
\end{itemize}

\section{Related Work}
\subsection{Semi-Supervised Domain Adaptation}
In many practical applications,  a few labeled data in the target domain are available. To consider labeled data in the target domain, the semi-supervised domain adaptation (SSDA) was proposed \cite{NIPS20104009,xiao2012semi,AAAI1714538,MME,kim2020attract,li2020online,qin2020opposite,BiAT}. Among these works, \cite{NIPS20104009} proposed a co-regulation based approach for SSDA, based on the notion of augmented space. \cite{xiao2012semi} proposed a kernel matching method mapping the labeled source data to the target data. Recently, \cite{MME} proposed a minimax entropy method and a new experiment setting with one or three labeled target data per class, which is viewed as the main baseline in our experiments. Following MME, \cite{qin2020opposite} applied an auxiliary classifier to improve target clustering and source scattering. Recently, meta-learning \cite{li2020online},  adversarial generation \cite{BiAT} and intra-domain discrepancy \cite{kim2020attract}  have been considered to improve SSDA task. Most of the former methods focus on explicit feature alignment for the unlabeled data between the source domain and the target domain. Furthermore, few of them consider how to make full use of labeled data from the target domain. In this paper, we propose an implicit feature alignment method without accessing source data while considering making full use of the labeled and unlabeled data in the target domain.  

\subsection{Hypothesis Transfer Learning}
In the setting of HTL,  we want to get a proper hypothesis (classifier) derived from the source domain and apply it to the target domain without accessing the source domain data \cite{HTL,feifei,kuzborskij2013n, SHOT}. Among existing works, \cite{feifei} imported the setting in one-shot object recognition problem and got satisfactory results. \cite{HTL} conducted a theoretical analysis of HTL built on the regression problem with domain shift. Recently, \cite{SHOT} applied the framework to the problem of unsupervised domain adaptation, via information maximization and clustering based pseudo labeling.  Specifically, \cite{SHOT} introduces a DeepCluster \cite{Deep_Cluster} based pseudo-labeling strategy that constructs centroid representation for each predicted class in the target domain and then obtains the pseudo labels by a nearest classifier. However, the strategy assumes that different classes in the target domain have similar intra-class diversity. On another side, the labeled data in the target domain can not be used effectively by this strategy.  Furthermore, most of the previous works focus on unsupervised domain adaptation, the methods using HTL for SSDA has not been fully researched  \cite{MM_HTL,Nelakurthi_Maciejewski_He_2018}.

\subsection{Semi-Supervised Learning}
Semi-Supervised Learning (SSL) aims to use the labeled and the unlabeled data together to facilitate the overall learning performance.  Here we list three classical methods: (1) consistency regularization, (2) entropy minimization, (3) data augmentation.
The consistency regularization methods include Mean Teacher \cite{tarvainen2017mean} and Virtual Adversarial Training \cite{VAT} e.t.,  which encourage the models to have stable outputs when applying perturbations to data. 
The entropy minimization methods encourage the model to make a confident prediction for unlabeled data. The methods consist of direct entropy minimization loss  \cite{EM} and indirect method via pseudo-labeling \cite{LP}.
Besides, the data augmentation methods like MixMatch \cite{NIPS20198749} also get a strong performance in semi-supervised tasks.  Although the above methods succeed, few of them consider the domain shift problem. \cite{Realistic_Evaluation_SSL} proposes that the performance of SSL methods can degrade drastically when dealing with domain shift problem. In our paper, we consider inter-domain as well as intra-domain alignments to boost the traditional semi-supervised methods for domain adaptation problem.

\section{Problem Formulation and Preliminaries}
Here we introduce the setting of SSDA. On the source domain, we have the labeled data $\mathbf{X}^{s} \in \mathbb{R}^{n\times d} $ and the corresponding labels $\mathbf{Y}^{s} \in \mathbb{R}^{n\times 1}$ . On target domain, we have a few labeled data $\mathbf{X}_{l}^{t} \in \mathbb{R}^{r\times d} $ with  corresponding labels $\mathbf{Y}_{l}^{t} \in \mathbb{R}^{r\times 1} $ , and the unlabeled data    $\mathbf{X}_{u}^{t} \in \mathbb{R}^{m\times d}$. Here the superscripts $s$  and $t$ denote the source domain and the target domain, respectively. $d$ denotes the dimension of data, $n$ ,$r$ and $m$ denote the number of the source domain data, the labeled target domain data and the unlabeled target domain data, respectively. The universal setting of SSDA is training a model to predict the labels of  $\mathbf{X}_{u}^{t} $, with the help of $\mathbf{X}^{s}$, $\mathbf{X}_{l}^{t}$ and their corresponding labels. In this paper, we firstly train a feature encoder $f_{\theta_{e}}$ followed by a bottleneck layer $f_{\theta_{b}}$, and a classifier $f_{\theta_{c}}$.  More specifically, we use cross entropy loss to train  $f_{\theta_{c}}$ ,$f_{\theta_{b}}$and $f_{\theta_{e}}$  on source domain, with a label smoothing operation to get more separated representations like \cite{SHOT}.  For the source data, we have the following loss function : 
\begin{equation}
\begin{array}{ll}
\mathcal{L}_{src}\left(\mathbf{X}^{s}, \mathbf{Y}^{s}; \theta \right)= 
-\mathbb{E}_{\left(x^{s}, y^{s}\right) \sim \mathbf{X}^{s},\mathbf{Y}^{s} } \sum_{k=1}^{K} y_{k}^{s} \log p_{k} \\
\\
y_{k}^{s}= \left\{\begin{array}{ll} 1-\varepsilon &  y_{k}^{s}=1 \\
\varepsilon /K &  otherwise.
\end{array}\right.
\end{array}
\label{eqa:source}
\end{equation}
 where  $\varepsilon$=0.1 is the smoothing parameter, $K$ is the number of classes, $p\left(y \mid x, \theta\right) = S \left(f_{\theta_{c}} 
\left(f_{\theta_{b}}
\left(f_{\theta_{e}}
\left(x\right)\right)\right)\right)$, $S(\cdot)$ is Softmax function.
 
Then on target domain, we freeze $\theta_{c}$ and update $\theta_{e},\theta_{b} $ using $\mathbf{X}_{l}^{t} $ , $\mathbf{Y}_{l}^{t} $  and $\mathbf{X}_{u}^{t} $ to predict the labels of  $\mathbf{X}_{u}^{t} $ . Our objective is to minimize  the following loss:

\begin{equation}
\begin{array}{ll}
\mathcal{L}_{tar}\left(\mathbf{X}^{t}_{l},\mathbf{X}^{t}_{u}, \mathbf{Y}^{t}_{l}; \theta\right)= 
\mathcal{L}_{lab} +\mathcal{L}_{reg}=\\
\\
-\mathbb{E}_{\left(x^{t}, y^{t}\right) \sim \mathbf{X}^{t}_{l},\mathbf{Y}^{t}_{l} } \sum_{k=1}^{K} y_{k}^{t} \log p_{k} + \mathcal{L}_{reg} ,
\end{array}
\label{eqa:target}
\end{equation}
where $y_{k}^{t}$ is the one-hot representation of label without label smoothing, and $\mathcal{L}_{reg}$ is the regularization losses. In the next, our aim is to design a proper $\mathcal{L}_{reg}$ on the target domain.

\begin{figure}[!hbtp]
\centering
\includegraphics[width=0.45\textwidth]{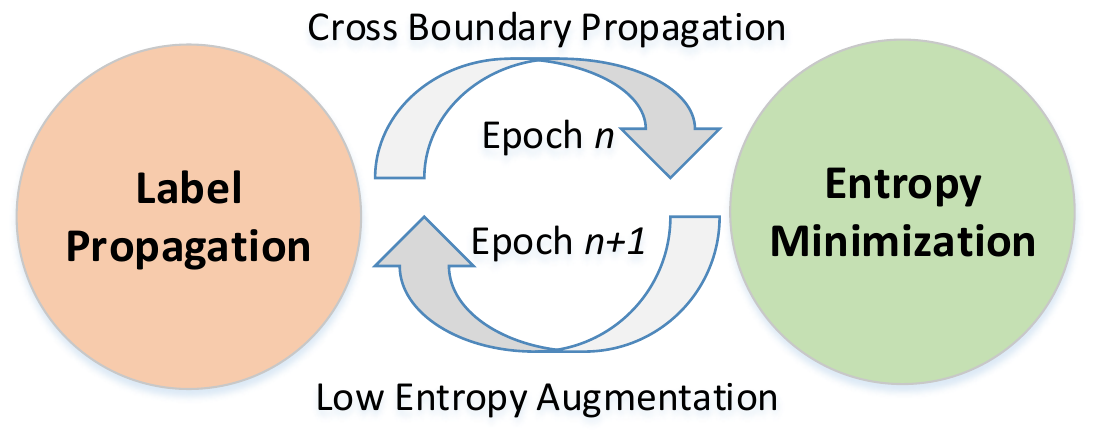}
\caption{The illustration of proposed mutual enhancement training process for entropy minimization and label propagation. At epoch $n$, the augmented label propagation is firstly performed to get the  pseudo-labels to correct the entropy minimization. At epoch $n+1$, some low entropy or low uncertainty examples which have experienced entropy minimization are utilized to augment the label propagation process. }

\label{fig:interaction}
\end{figure} 

\begin{figure*}[!hbtp]
\centering
\centering
\includegraphics[]{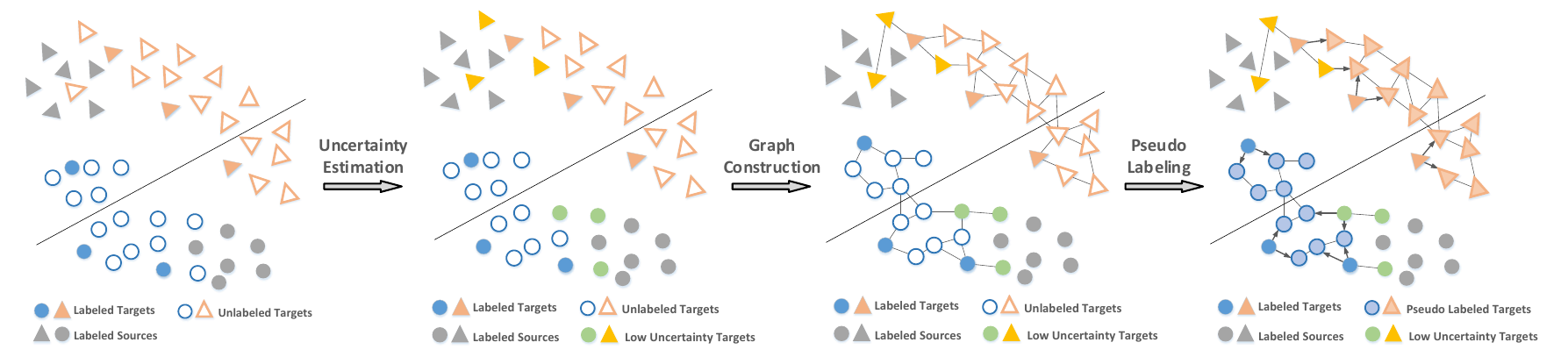}
\caption{Illustration of the augmented label propagation process. The first figure demonstrates the representations of domain data using the pre-trained model from the source domain. Note that the source data are unseen in the label propagation process and the black line denotes the fixed classify boundary trained on the source domain. The second figure shows the situation where the low uncertainty target data are selected. The last two figures demonstrate the graph construction and label propagation process. }

\label{fig:train_process}
\end{figure*} 
\section{The Proposed SSDA Method}
In this section, we discuss how to design $L_{reg}$ on the target domain. Firstly, we revisit entropy minimization from the view of pseudo labeling strategy and explore its advantage/limitation for the unlabeled target data. Then, a pseudo labeling method derived from label propagation is applied for alleviating the limitation of entropy minimization.  In the proposed pseudo labeling method, some low entropy examples which have experienced entropy minimization are utilized to augment the propagation process.  As a result, we build a mutual enhancement algorithm in which the entropy minimization and label propagation promote each other. Figure \ref{fig:interaction} demonstrates the enhancement process. Finally, a Virtual Adversarial Training (VAT \cite{VAT}) method and a class-diversity regularizer are utilized to enhance the smoothness in propagation process and alleviate the trivial prediction at the beginning of adaptation. 

\subsection{Mutual Enhancement Training}
Missing the explicit representation alignment on both domains, the traditional domain adaptation methods do not fit our HTL setting. However, to make full use of the unlabeled data, applying Entropy Minimization \cite{EM} on the unlabeled data is beneficial for getting more discriminative representations by forcing the representations of  the unlabeled data to be away from decision boundary \cite{Zhang2019CVPR,SHOT}.  Following with \cite{EM}, the entropy minimization loss can be defined as:

\begin{equation}
\mathcal{L}_{ent}\left(\mathbf{X}^{t}_{u}; \theta \right)= 
-\mathbb{E}_{ x^{t} \sim \mathbf{X}^{t}_{l} } \sum_{k=1}^{K}  
p_{k}  \log p_{k}
\label{eqa:ent}
\end{equation}

When using the entropy minimization, we consider a hard situation where the representations of unlabeled data span the decision boundary. As demonstrated by the first figure in Figure \ref{fig:train_process}, the fixed classification boundary divides the class into one side that the label can be predicted rightly, and the other side that the classifier will have the wrong prediction.  This will lead to two different results:  the model will be more confident with the correct classification, but also more confident with the wrong classification. To alleviate the limitation of entropy minimization, Equation \ref{eqa:ent} motivates us to understand it from the the view of pseudo labeling.  The  Equation \ref{eqa:ent} encourages the largest probability elements to be closed to $1$, hence its effect can be approximated by:

\begin{equation}
\mathcal{\hat{L}}\left(\mathbf{X}^{t}_{u}; \theta \right)= 
-\mathbb{E}_{ x^{t} \sim \mathbf{X}^{t}_{l} } \sum_{k=1}^{K}  
q_{k}  \log p_{k},
\label{eqa:enthat}
\end{equation}
where $q$ is the pseudo label and can be defined as:
\begin{equation}
q_{k}=\left\{\begin{array}{ll}
1, & \text { if } k = argmax_{i}(p_{i}) \\
0, & \text { otherwise }
\end{array}\right.\end{equation}

Therefore, based on entropy minimization, we consider how to get more accurate pseudo-labels for the unlabeled data $\mathbf{X}^{t}_{u}$. We assume that the representations of the target domain data exist in a smooth manifold space, where a nearest neighbor graph can be constructed by incorporating the labeled target data and the unlabeled target data. Our solution is to use Label Propagation \cite{LP} to propagate the label information from the labeled data to the unlabeled, or propagate soft label from the low uncertainty unlabeled data to the high uncertainty ones. Concretely, if $r$ denotes the number of labeled target data $\mathbf{X}^{t}_{u}$, and $m$ denotes the total number of low uncertainty data and high uncertainty data in $\mathbf{X}^{t}_{u}$,  we can get an similarity matrix $W \in \mathbb{R}^{(r+m)\times(r+m)}$ :

\begin{equation}
W_{i j}=\exp \left(sim \left( x_{i}, x_{j}\right) \right),
\label{similarity}
\end{equation}
where $W_{ij}$ denotes the similarity of data $x_{i}$  and data $x_j$, $sim(\cdot,\cdot)$ denotes the cosine similarity. To describe the $\hat{k}$ nearest graph using matrix $W$, we preserve the largest $\hat{k}$  elements of each row and set the others as zero.
Then we can get probability matrix $Z \in \mathbf{R}^{(r+m)\times K} $ by closed solution  \cite{LPnips}: 
\begin{equation}
Z=(I-\alpha {W})^{-1} Y,
\end{equation}where $I$ is the identity matrix, $\alpha=0.9$ is the fixed weight of $W$, and the label matrix $Y \in \mathbb{R}^{(r+m)\times K}$is defined as:
\begin{equation}Y_{i j}:=\left\{\begin{array}{ll}
1, & \text { if } x_i \in X_{l}^{t} \cup X_{lu}^{t}  \quad  and \quad  y_{i}=j \\
0, & \text { otherwise },
\end{array}\right.\end{equation}where $X_{lu}^{t}$ is the low uncertainty data estimated from $\mathbf{X}^{t}_{u}$. For simplicity, we select one lowest uncertainty example according its predicted soft label for each class. For example, the total number of low uncertainty data is $K$ for a $K$ class dataset.
Finally we get the pseudo labels $\hat{y}_{i}^{t} $ and classification loss as: \begin{equation}
\hat{y}_{i}^{t}=\arg \max _{j} z_{i j}
\label{eqa:psz}
\end{equation}

\begin{equation}
\mathcal{L}_{ps} \left(\mathbf{X}^{t}_{u},\mathbf{Y}^{t}_{u}; \theta\right)= -\mathbb{E}_{\left(x^{t}, y^{t}\right) \sim \mathbf{X}^{t}_{u},\mathbf{Y}^{t}_{u} } \sum_{k=1}^{K} \hat{y}_{k}^{t} \log p_{k}
\label{eqa:ps}
\end{equation}
Figure \ref{fig:train_process} illustrates the augmented label propagation spanning the fixed classification boundary and getting more accurate pseudo-labels for the unlabeled target data.  

To make the entropy minimization and label propagation promote each other, we build a mutual enhancement training method as following steps (see Figure \ref{fig:interaction}):  (1) \textbf{Label Propagation to Entropy Minimization}, at the beginning of each training epoch, the augmented label propagation is firstly performed to get the more  accurate pseudo-labels; then entropy minimization loss $\mathcal{L}_{ent}$ is optimized with pseudo loss $\mathcal{L}_{ps}$ to correct the gradient direction for $\mathcal{L}_{ent}$. (2)\textbf{Entropy Minimization to Label Propagation}, at the beginning of next epoch, some low entropy examples which have experienced entropy minimization are utilized to augment the label propagation process.  The reason behind this efficient training method is that the 
transferred model from the source domain provides good initialization for target task. Moreover, as demonstrated by the second figure in Figure \ref{fig:framework} , the low uncertainty target data derived from source model are more closer to source data compared to the high uncertainty data. Therefore, the representations of low uncertainty target data have better alignment with the source representations, hence improve the whole domain alignment via augmented label propagation.
\subsection{Local Smoothing}
As demonstrated by Equation \ref{similarity}, label propagation relies on the following prior assumption \cite{LPnips}:  Nearby data are likely to have the same label. This needs the representations to be sufficiently smooth for the data points and their neighbors. \cite{LPnips} called this smooth as local consistency or local smoothness; \cite{VAT,BiAT} imposed perturbations on data points via Virtual Adversarial Training (VAT) to enhance the local smoothness of representations. Following with \cite{VAT}, we use $x^{t}$ to denote the labeled and unlabeled data from target domain, then the objective of VAT can be defined as follows:

\begin{equation}
\label{eqa:vadv}
\mathcal{L}_{\mathrm{vadv}}\left(x^{t}, \theta\right)=D\left[p\left(y \mid x^{t}, \theta \right), p\left(y \mid x^{t}+r_{\mathrm{vadv}}, \theta\right)\right],
\end{equation}where \begin{equation}
r_{\mathrm{vadv}}:=\underset{r ;\|r\| \leq \epsilon}{\arg \max } D\left[p\left(y \mid x^{t}, \theta \right), p\left(y \mid x^{t}+r\right),\theta \right],
\end{equation}and  $p\left(y \mid x_{t}, \theta\right)$ is the current estimate before adding perturbation, $\epsilon$ is the threshold for perturbation value $r_{vadv}$, $D[\cdot]$ is the Kullback–Leibler divergence (KLD, \cite{6832827}).

\subsection{Class Balancing}
Finally, to reduce the risk where most of the unlabeled data are assigned  the same pseudo labels at the beginning of adaptation. We want a regularization to encourage the number of each class examples to be equal. \cite{Self_Labeling} explored a similar limitations in unsupervised image classification tasks via the Optimal Transport framework \cite{8641476,7586038}. \cite{SHOT} also imports the regularization item into unsupervised domain adaptation as follows:
\begin{equation}
\label{eqa:div}
\mathcal{L}_{d i v}\left(\mathbf{X}^{t}_{u}; \theta\right)=\sum_{k=1}^{K} \hat{p}_{k} \log \hat{p}_{k},\end{equation}
where \begin{equation}\hat{p}_{k}=\mathbb{E}_{x^{t} \sim \mathbf{X}_{u}^{t}}\left[p_{k} \right]
\end{equation}
\subsection{Overall Objective of Regularization}
To summarize, the  overall objective on the unlabeled target data is:
\begin{equation}
\mathcal{L}_{reg} = \lambda_{0}\mathcal{L}_{ps}+\mathcal{L}_{ent}+ +\mathcal{L}_{vadv}+\mathcal{L}_{div},
\label{eqa:overall}
\end{equation} where $ \lambda_{0} $ is the weight of $\mathcal{L}_{ps}$. Note that both  $\mathcal{L}_{ps} $ and $\mathcal{L}_{ent}$ are treated as pseudo labeling losses, hence a balancing term $ \lambda_{0} $ is introduced. Algorithm \ref{alg:algorithm1} demonstrates proposed adaptation method.
\begin{algorithm}[tb]
\caption{Adaptation Stage of The Proposed Method}
\label{alg:algorithm1}
\begin{flushleft}
\textbf{Input}: The  labeled target data $\mathbf{X}_{l}^{t} $ , $\mathbf{Y}_{l}^{t}$ and the unlabeled target data $\mathbf{X}_{u}^{t}$ \\
\textbf{Parameter}: Encoder parameters $\theta_{e}$, bottleneck layer $\theta_{b}$, classifier parameters $\theta_{c}$, learning rates $ \alpha_{0}, \alpha_{1}$ \\
\textbf{Output}: the updated parameters $\theta_{e}$, and $\theta_{b}$
\end{flushleft}
\begin{algorithmic}[1] 
\STATE Freeze $\theta_{c} $
\WHILE{not convergence}
\STATE Compute pseudo labels for $\mathbf{X}_{u}^{t}$ via Equation \ref{eqa:psz}
\FOR {$step = 0 \to T$ }
\STATE Sample batch data $\mathbf{x}_{l}^{t}, \mathbf{y}_{l}^{t},\mathbf{x}_{u}^{t}$
\STATE Compute $\mathcal{L}_{lab}$ using $\mathbf{x}_{l}^{t}, \mathbf{y}_{l}^{t}$ via Equation \ref{eqa:target}
\STATE Compute $\mathcal{L}_{ent}$ using $\mathbf{x}_{u}^{t}$ via Equation \ref{eqa:ent}
\STATE Compute $\mathcal{L}_{ps}$ using $\mathbf{x}_{l}^{t}, \mathbf{y}_{l}^{t},\mathbf{x}_{u}^{t}$ ,via Equation \ref{eqa:ps}
\STATE Compute $\mathcal{L}_{vadv}$ using $\mathbf{x}_{l}^{t},\mathbf{x}_{u}^{t}$ via Equation \ref{eqa:vadv}
\STATE Compute $\mathcal{L}_{div}$ using $\mathbf{x}_{l}^{t}, \mathbf{y}_{l}^{t}$ via Equation \ref{eqa:div}
\STATE Evaluate  $\nabla_{\theta_{e}} \mathcal{L}_{\mathcal{T}_{tar}}$ and $ \nabla_{\theta_{b}} \mathcal{L}_{\mathcal{T}_{tar}}$ via Equation \ref{eqa:target}.
\STATE Update $\theta_{e} $ : $\theta_{e}^{\prime} \leftarrow \theta_{e}- \alpha_{0} \nabla_{\theta_{e}} \mathcal{L}_{\mathcal{T}_{tar}}$
\STATE Update $\theta_{b} $ : $\theta_{b}^{\prime} \leftarrow \theta_{b}- \alpha_{1} \nabla_{\theta_{b}} \mathcal{L}_{\mathcal{T}_{tar}}$
\ENDFOR
\ENDWHILE
\end{algorithmic}
\end{algorithm}

\section{Experiments}

\subsection{Datasets}
We adopt three public datasets including \textbf{Office-31} \cite{office}, \textbf{Office-Home} \cite{venkateswara2017Deep} and \textbf{DomainNet} \cite{peng2018moment}.

 \textbf{Office-31}  is a small size dataset containing 3 domains (Amazon, Webcam and Dslr) and 31 categories in each domain.  The Amazon domain contains images from the Amazon website, and the Webcam and Dslr domain contain the office environment images taken with varying lighting and pose changes using a webcam and dslr camera, respectively. 

 \textbf{Office-Home} is a middle size dataset and was created to evaluate domain adaptation algorithms for object recognition. It consists of images from 4 domains: Art, Clipart, Product and Real-World images. For each domain, the dataset contains images of 65 object categories built typically from office and home environments. 

 \textbf{DomainNet } contains six domains including Clipart, Infograph, Painting, Real, Quickdraw, with each domain containing 345 categories of common objects.  Like \cite{MME}, we pick 4 domains (Real, Clipart, Painting, Sketch), and 126 classes for each domain. Furthermore, we also focus on the adaptation scenarios where the target domain is not real images, and perform adaptation experiments on 7 scenarios from the 4 domains.


\subsection{Baselines}
The baselines include four categories:
\textbf{None}, None Adaptation ( S+T \cite{MME}). S+T is a method training the model with labeled source and labeled target data without using unlabeled data from the target domain.  

\textbf{SSL}, Semi-Supervised Learning ( like ENT \cite{EM,MME}). ENT is a semi-supervised method that minimizes the entropy of the unlabeled target data.

\textbf{UDA}, Unsupervised Domain Adaptation (such as  DANN \cite{ganin2015unsupervised}, ADR \cite{saito2018adversarial}, CDAN \cite{long2018conditional}, \cite{SHOT}). DANN explored domain-invariant features using gradient reversal. ADR proposed Adversarial Dropout Regularization for adversarial generation methods, to encourage the generator to output more discriminative features for the target domain. CDAN proposed two conditioning strategies consisting of multi-linear conditioning and entropy conditioning to guarantee the transferability. SHOT proposed a privacy preserving method for unsupervised domain adaptation via hypothesis transfer. For the unsupervised domain adaptation  methods (DANN, ADR, and CDAN, SHOT), the labeled target data were put with the source domain data during the training process.

\textbf{SSDA}, Semi-Supervised Domain Adaptation ( including MME \cite{MME}, BiAT \cite{BiAT}, APE \cite{kim2020attract}, M-MME \cite{li2020online}, MESH-nA). MME proposed minimax entropy loss for the SSDA problem.  Based on MME, M-MME introduce meta learning paradigm \cite{pmlr-v70-finn17a} to improve generalization. BiAT uses bidirectional adversarial training for generating samples between the source and target domain. APE solves the intra-domain discrepancy problem with Attracting, Perturbing, and Exploring. We reproduced the experiments based on the origin codes of APE. {MESH-nA} denotes our method with non-{A}ugmentated label propagation.

\subsection{Settings}

\subsubsection{Data Preparation. }  When training the hypothesis (classifier module), we split the data in the source domain into the training set and the validation set. Specifically, on Office-31 and Office-Home, we split the training set and the validation set as 0.9 : 0.1; on DomainNet, the partition is set as 0.98 : 0.02. For source domain data, we use the standard data augmentation methods consisting of random horizontal flip, random crop, and data normalization. For a fair comparison with baselines, we use the same partition for the target domain as \cite{MME}.

\subsubsection{Implementation Details}
\textbf{ Module setting.} We select three encoder backbones including AlexNet \cite{krizhevsky2012imagenet}, VGGNet \cite{simonyan2014very} and ResNet-34 \cite{he2016deep}. For the AlexNet and the VGGNet, we add a bottleneck layer after the last layer of the encoder. Then we use a classifier with one normalized fully connected layer. For the ResNet-34, we drop the last layer of the model and add a  bottleneck layer like the former backbones, and use a new classifier with one fully connected layer.   

\textbf{Uncertainty estimation. } To estimate the low uncertainty examples, we use MC-Dropout \cite{pmlr-v48-gal16}, which opens the dropout operation and calculates the mean of repeated outputs.  We use the entropy (see Equation \ref{eqa:ent}) to denote the uncertainty of an example. For all the datasets, the number of low uncertainty examples is set to 1 for each class.

\textbf{Hyper-parameter setting. } We set  $ \lambda_{0}=0.5 $ for Office-31 and   $ \lambda_{0}=0.3$ for Office-Home and  DomainNet according to their validation sets. The batch size is set to $\{64,32,64\}$ for Office-31, Office-Home and DomainNet. For all experiments, $\hat{k} $ is set to 10,  the  learning rates  for $\{\theta_{e}, \theta_{b}, \theta_{c}\}$ are set to \{0.001,0.01,0.01\}. We use three seeds $\{2021,2022,2023\}$ and  \textbf{repeat three times} to report the mean results.   Besides, our experiments are implemented with Pytorch \footnote{https://pytorch.org/} and running on one RTX 3090 GPU.


\begin{table}[!bhtp]
\centering
\caption{Semi-supervised domain adaptation accuracies(\%) on Office-31 using AlexNet. N-shot denotes n labeled targets for each class. The capitals D, A and W denote Dslr, Amazon, Webcam, respectively. Ave denotes the average performance on the n-shot task. Due to space limit, only the mean results are reported like \cite{MME}.}

\begin{tabular}{c|ccc|ccc}
\toprule
\multicolumn{1}{l}{} & \multicolumn{3}{c}{1-shot}                                                              & \multicolumn{3}{c}{3-shot}                                                                                \\ \hline
Methods              & \multicolumn{1}{c}{D-A}  & \multicolumn{1}{c}{W-A}  & Ave                              & D-A                               & W-A                               & Ave                              \\ \hline
S+T                  & \multicolumn{1}{c}{50.0} & \multicolumn{1}{c}{50.4} & 50.2                              & 62.4                              & 61.2                              & 61.8                              \\
DANN                 & 54.5                     & 57.0                     & 55.8                              & 65.2                              & 64.4                              & 64.8                              \\
ADR                  & 50.9                     & 50.2                     & 50.6                              & 61.2                              & 61.4                              & 61.3                              \\
CDAN                 & 48.5                     & 50.4                     & 49.5                              & 60.3                              & 61.4                              & 60.8                              \\
SHOT                 & \multicolumn{1}{c}{52.9} & \multicolumn{1}{c}{50.4} & 51.7                              & 65.4                              & 65.3                              & 63.3                              \\
ENT                  & 50.0                     & 50.7                     & 50.4                              & 64.0                              & 66.2                              & 65.6                              \\
MME                  & 55.8                     & 57.2                     & 56.5                              & 67.8                              & 67.3                              & 67.5                              \\
BiAT                 & \multicolumn{1}{c}{54.6} & \multicolumn{1}{c}{57.9} & 56.3                              & 68.5                              & 68.2                              & 68.3                              \\
APE     & \multicolumn{1}{c}{-} & \multicolumn{1}{c}{-} & -                              & 67.6                              & 69.0                              & 68.3                              \\
\midrule
\textbf{MESH-nA }          & \multicolumn{1}{c}{55.5} & \multicolumn{1}{c}{58.8} & 57.2                              & 68.5                              & 68.6                              & 68.6                              \\
\textbf{MESH }               & \textbf{75.7}            & \textbf{76.6}            & {\textbf{76.2}} & {\textbf{76.8}} & {\textbf{78.5}} & {\textbf{77.7}}\\
\bottomrule
\end{tabular}
\label{tab:office}
\end{table}

\begin{table*}[!h]
\centering
\caption{Semi-supervised domain adaptation results on Office-Home dataset using VGGNet. The capitals denote the following domains: P (Product), C (Clipart), A (Art), R (Real World). N-shot denotes n labeled targets for each class. Due to space limit, only the mean results are reported like \cite{MME}.}

\begin{tabular}{c|c|ccccccccccccc} 
\toprule
\multicolumn{15}{c}{1-shot} \\ \hline
 Categories& Methods&P-C & P-A &C-R &A-P &R-C & R-P&	R-A	&P-R &	A-C&	A-R	&	C-A&	C-P	&Ave \\
\midrule

\multicolumn{1}{c|}{None}                  & \multicolumn{1}{c|}{S+T}        & 37.0          & 52.0          & 64.5          & 63.6          & 39.5          & 75.3          & 61.2          & 71.6          & 37.5          & 69.5          & 51.4          & 65.9          & 57.4          \\ \midrule
\multicolumn{1}{c|}{\multirow{4}{*}{UDA}}  & \multicolumn{1}{c|}{DANN}       & 45.9          & 51.3          & 64.2          & 64.3          & 52.0          & 75.7          & 62.7          & 72.7          & 44.4          & 68.9          & 52.3          & 65.3          & 60.0          \\
\multicolumn{1}{c|}{}                      & \multicolumn{1}{c|}{ADR}        & 47.8          & 51.4          & 64.8          & 63.9          & 39.7          & 76.2          & 60.2          & 71.8          & 39.0          & 68.7          & 50.0          & 65.2          & 63.0          \\
\multicolumn{1}{c|}{}                      & \multicolumn{1}{c|}{CDAN}       & 37.2          & 44.5          & 58.7          & 67.7          & 43.3          & 75.7          & 60.9          & 69.6          & 39.8          & 64.8          & 41.6          & 66.2          & 57.4          \\
\multicolumn{1}{c|}{}                      & \multicolumn{1}{c|}{SHOT}       & 38.9          & 51.9          & 68.8          & 66.2          & 44.5          & 75.9          & 61.5          & 73.9          & 42.6          & 71.9          & 56.5          & 65.7          & 59.9          \\ \midrule
\multicolumn{1}{c|}{SSL}                   & \multicolumn{1}{c|}{ENT}        & 21.3          & 44.6          & 62.1          & 66.0          & 23.7          & 77.5          & 64.0          & 74.6          & 22.4          & 70.6          & 25.1          & 67.7          & 51.6          \\ \midrule
\multicolumn{1}{c|}{\multirow{3}{*}{SSDA}} & \multicolumn{1}{c|}{MME}        & 46.2          & 56.0          & 68.0          & 68.6          & 49.1          & 78.7          & 65.1          & 74.4          & 45.8          & 72.2          & 57.5          & 71.3          & 62.7          \\
\multicolumn{1}{c|}{}                      & \multicolumn{1}{c|}{\textbf{MESH-nA}} & 42.8          & 59.0          & 63.6          & 74.5          & 47.0          & 77.8          & 63.8          & 73.7          & 42.9          & 70.6          & 58.4          & 72.2          & 62.2          \\
\multicolumn{1}{c|}{}                      &\textbf{MESH  }                         & \textbf{62.1} & \textbf{65.4} & \textbf{74.6} & \textbf{76.5} & \textbf{62.5} & \textbf{82.1} & \textbf{68.5} & \textbf{79.0} & \textbf{60.3} & \textbf{76.1} & \textbf{63.3} & \textbf{75.4} & \textbf{70.5} \\ 
 \midrule \midrule
 \multicolumn{15}{c}{3-shot} \\ \hline
 Categories& Methods&P-C & P-A &C-R &A-P &R-C & R-P&	R-A	&P-R &	A-C&	A-R	&	C-A&	C-P	&Ave \\
\midrule

\multirow{1}{*}{None} & S+T  &47.2 & 55.9 &69.7 &69.4 &49.6 & 78.6&	63.6	&72.7 &	47.5&	73.4	&	56.2&	70.4&62.9 \\
\midrule
\multirow{4}{*}{UDA} &DANN&52.4 & 56.3 &68.7 &69.5 &56.1 & 77.9&	63.7	&73.6 &	50.0&	72.3	&	56.4&	69.8	&63.9 \\
&ADR &47.8 & 55.8&69.3 &69.9 &49.0 & 78.1&	62.8	&73.6 &	49.3&	73.3	&	56.3&	71.4&63.0 \\
&CDAN  &45.1 &50.3 &65.9 &74.7 &50.2 & 80.9&	62.1	&70.8 &	46.0&	71.4	&	52.9&	71.2&61.8 \\
&SHOT  &49.3 & 58.5 &73.3 &76.3 &51.4 & 82.4&	65.2	&76.5 &	50.6&	75.4	&	60.4&	74.9	&66.2 \\
\midrule
\multirow{1}{*}{SSL}&ENT &46.8 & 56.9 &72.9 &73.0 &48.3 & 81.6&	65.5	&76.6 &	44.8&	75.3	&	59.1&	77.0	&64.8 \\
\midrule
\multirow{3}{*}{SSDA}&MME  &53.1 & 59.2 &72.9 &75.7&56.9 &82.9&	65.7	&76.7 &	54.9&	75.3	&	61.1&	76.3&67.6 \\
 &\textbf{MESH-nA} &55.8 & 61.7 &75.1 &77.9 &54.1 & 81.9&	66.4	&76.0 &	54.5&	76.5	&	61.4&	\textbf{78.9}	&68.3 \\
&\textbf{MESH} &\textbf{66.3} &\textbf{ 64.0 }&\textbf{75.6} &\textbf{80.1} &\textbf{69.3} & \textbf{86.6}&	\textbf{69.8}	&\textbf{79.3} &	\textbf{64.0}&	\textbf{77.8}	&	\textbf{63.7}&	78.3	&\textbf{72.9} \\
 
\bottomrule
\end{tabular}

\label{tab:officehome}
\end{table*}
\begin{table*}[!tbhp]
\centering
\caption{Domain adaptation results on DomainNet dataset with ResNet34. N-shot denotes n labeled targets for each class. The capitals R, S, P, C denote the following domains: R: real, S: sketch, P : painting, C: clipart. Due to space limit, only the mean results are reported like \cite{MME}.}
\begin{tabular}{c|c|cccccccc|cccccccc} 
\toprule
& & &\multicolumn{8}{c}{1-shot}  &\multicolumn{7}{c}{3-shot} \\  \midrule
Categories &Methods &R-S  &R-P &C-S  &R-C 	 &P-C	&P-R &S-P 	&Ave &R-S  &R-P &C-S  &R-C 	 &P-C	&P-R &S-P 	&Ave\\ \midrule
\multirow{1}{*}{None} & S+T  &46.3   & 60.6 &50.8    &55.6    &56.8       	 &71.8		 &56.0 	   &56.9  &50.1   &62.2   &55.0  &60.0 	    &59.4	     &73.9    &59.5 	   &60.0\\ \midrule
\multirow{4}{*}{UDA} &DANN  &52.2   &61.4  &52.8	   & 58.2    &56.3     	  &70.3		  &57.4 	&58.4 &54.9   &62.8    &55.4  &59.8 	   &59.6	    &72.2    &59.9 	      &60.7\\
&ADR      &49.0   &61.3  &51.0 	  &  57.1   &57.0         &72.0		 &56.0 	   &57.6 &51.1   &61.9    &54.4  &60.7 	  &60.7	       &74.2    &59.9        &60.4\\ 
&CDAN  &54.5   &64.9  &53.1	   &65.0      &63.7        &73.2	  &63.4 	&62.5 &59.0   &67.3   &57.8  &69.0 	   &68.4	     &78.5    &65.3 	   &66.5\\
&SHOT & 60.4 & 67.0 & 61.0 & 69.0 & 69.4 & 79.4 & 62.4 &67.0 &59.3  &66.5 &61.2  &68.7 	 &69.3	&80.0 &63.4 	&66.9\\ \midrule
\multirow{1}{*}{SSL} &ENT  &52.1   &65.9  &54.6    &65.2 &65.4        &75.0		 &59.7     &62.6 &61.1    &69.2    &60.0  &71.0	   &71.1	    &78.6   &62.1 	    &67.6\\ \midrule
\multirow{6}{*}{SSDA}  
&MME    &61.0   &67.7  &56.3    &70.0	   &69.0     	 &76.1	   &64.8 	&66.4 &61.9    &69.7    &61.8  &72.2      &71.7	      &78.5     &66.8 	     &68.9\\
&BiAT     &58.5  &68.0 &57.9  &73.0  &71.6	 &77.0	 &63.9 	&67.1 &62.1    &68.8    &61.5  &74.9   &74.6	      &78.6   &67.5 	 &69.7\\ 
&M-MME    &-  &-  &-    &-	   &-     	 &-	   &- 	&- 
& 63.8 &  70.3 & 62.8 & 73.5 &72.8 & 79.2 & 68.0 &70.1 \\
&APE   & 61.7 &67.2 & 57.4 & 67.2  & 68.5 & 75.7 & 60.4 & 65.4 
      & 66.2 &71.2 & 63.9 & 72.9  & 72.7 & 78.0 & 67.9 & 70.4  \\
&\textbf{MESH-nA}     &64.5  &67.3 &60.2  &71.4 	 &70.9	&73.9 &64.5 &67.5 
&65.2  &69.6  &64.3  &73.9 	&74.6	   &77.1      &66.0 	  &70.2\\

&\textbf{MESH}  &\textbf{76.7}  &\textbf{78.1} &\textbf{71.9}  &\textbf{79.7 }	 &\textbf{77.0}	&\textbf{85.4} &\textbf{77.7} 	&\textbf{78.1}
&\textbf{75.9}  &\textbf{79.0}  &\textbf{74.9}  &\textbf{80.5} 	&\textbf{77.5}	   &\textbf{88.5}     &\textbf{80.0} 	  &\textbf{79.5}\\

\bottomrule
\end{tabular}

\label{tab:domainnet}
\end{table*}

\subsection{Results}
For the n-shot (n labeled examples per target class) domain adaptation task with AlexNet on \textbf{Office-31}, we conduct two challenging domain adaptation scenarios including Dslr to Amazon and Webcam to Amazon (see Table \ref{tab:office}). Our method gets competitive performance compared with recent SSDA methods such as MME and BiAT, while achieving the privacy preserving for the source domain data. For example, our method gets 75.7\% accuracy on 1-shot Dslr to Amazon, which outperforms the MME method nearly 20\%.

For \textbf{Office-Home}, we use VGGNet as a backbone and perform n-shot domain adaptation experiments. From Table \ref{tab:officehome}, we found that our method outperforms all the methods with a large margin. We also compare our method with BiAT and APE with AlexNet backbone. The average performance (58.5\%) of our method on twelve 3-shot tasks outperforms BiAT (56.4\%) and APE (55.6\%).

For the large dataset \textbf{DomainNet}, we use ResNet34 as a backbone and perform n-shot domain adaptation experiments. Like the previous work in SSDA, we adopt 7 domain adaptation scenarios. Surprisingly, compared to the recent methods, we still get competitive adaptation performance on the target domain without accessing the source domain data. 

\subsection{Detail Analyses}
In this section, we provide deeper analyses toward the following problems:  1) What will happen when the mutual enhancement training is broken? 2) What is the effect of each regularization term? 3) Is our method still working when given various hyper-parameters?

\begin{figure*}[!thp]
\centering
\subfigure[\scriptsize{accuracy on low uncertainty data}]{
\label{fig:data1}
\centering
\includegraphics[width=0.23 \textwidth]{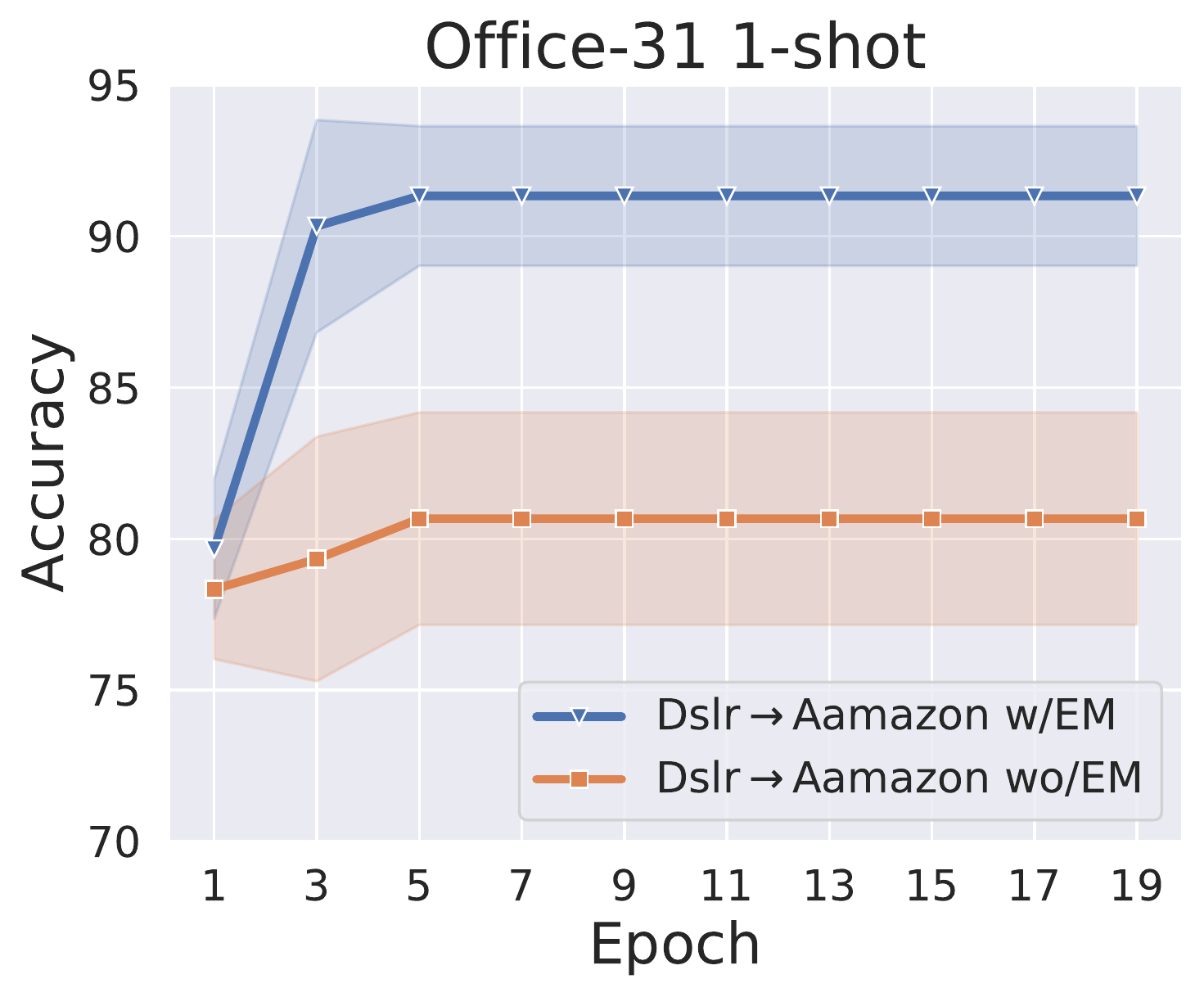}
}
\subfigure[\scriptsize{accuracy on low uncertainty data}]{
\label{fig:data2}
\centering
\includegraphics[width=0.23 \textwidth]{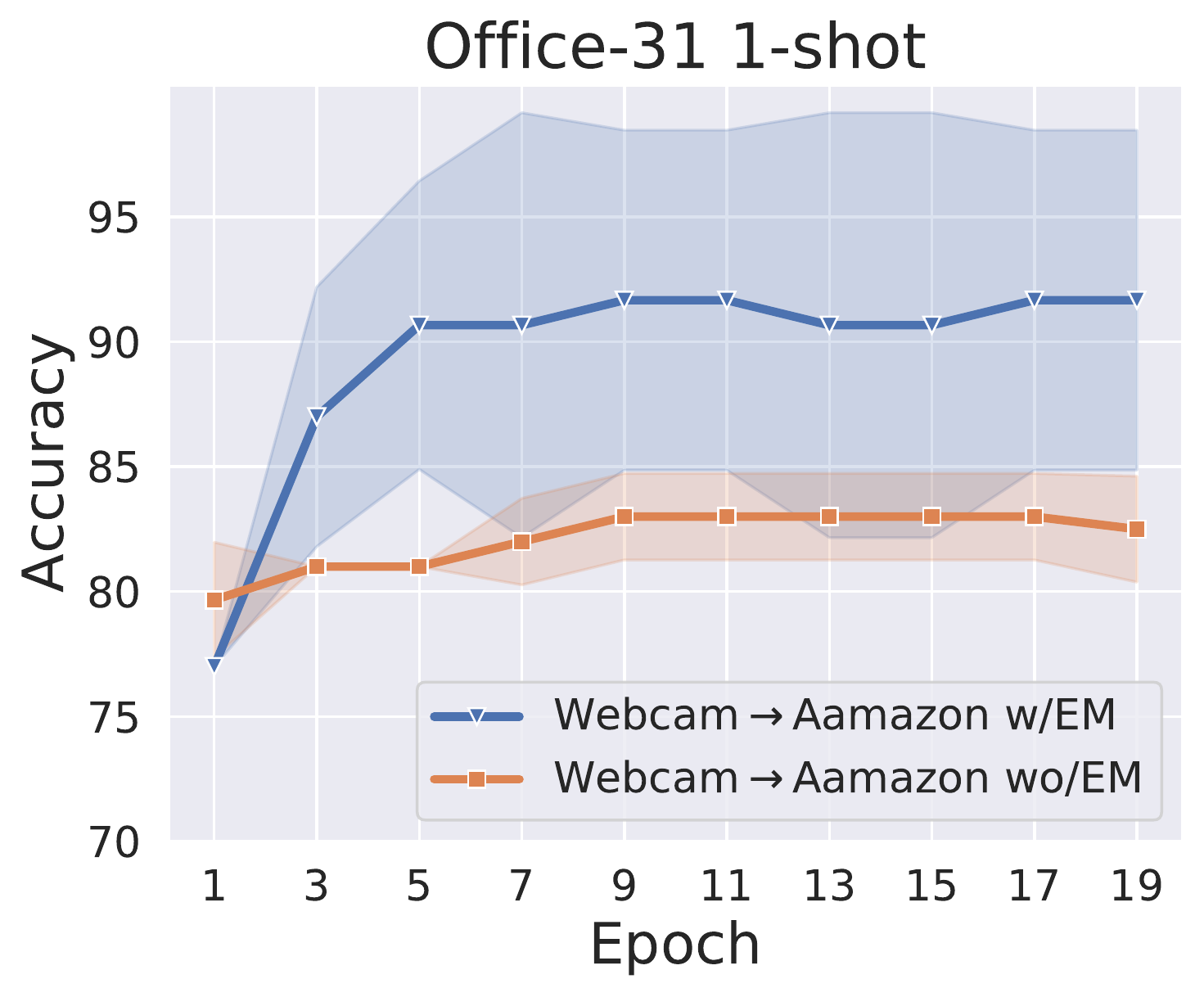}
}
\subfigure[\scriptsize{different augmentation for 1-shot}]{
\label{fig:data3}
\centering
\includegraphics[width=0.23 \textwidth]{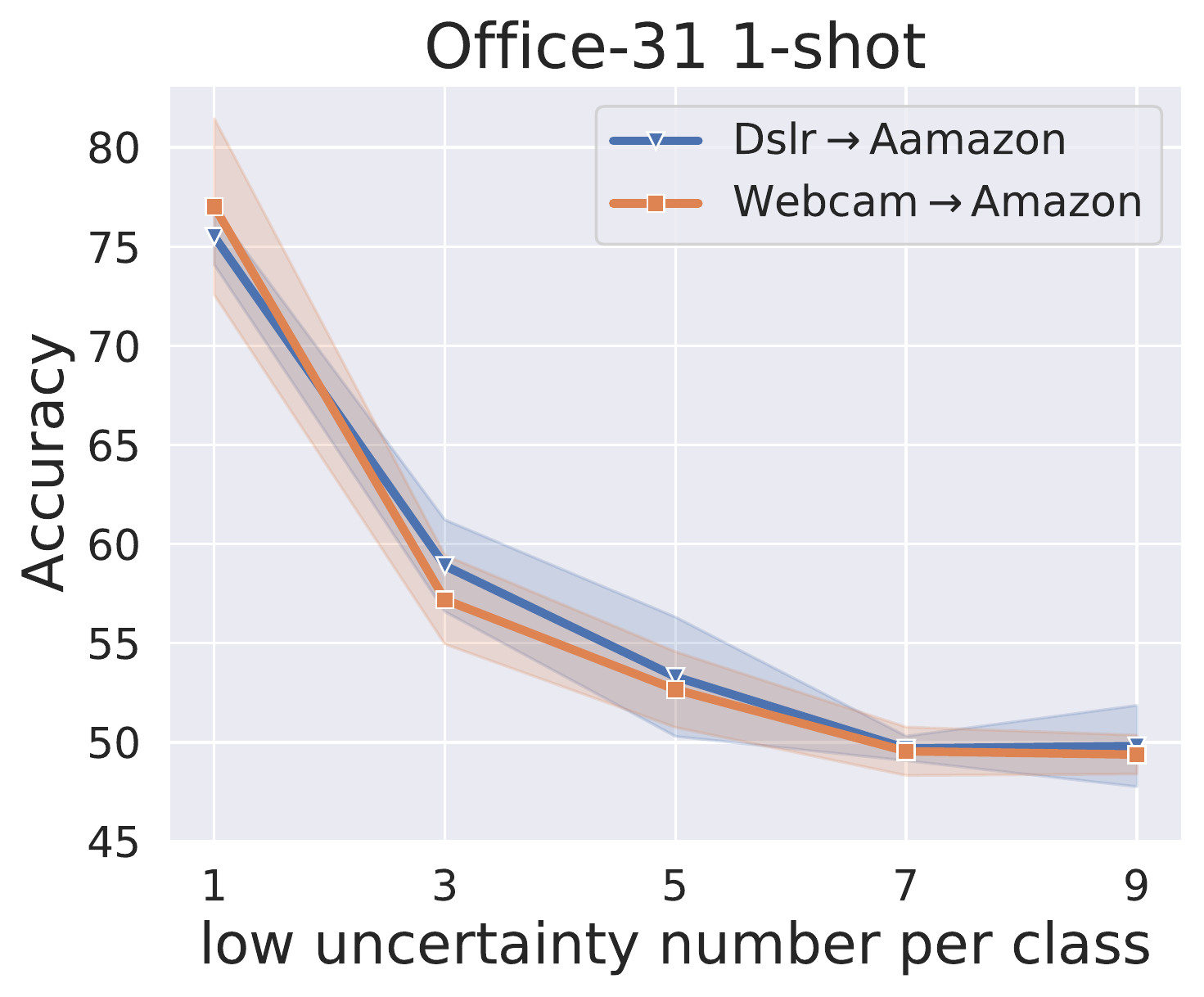}
}
\subfigure[\scriptsize{different augmentation for 3-shot}]{
\label{fig:data4}
\centering
\includegraphics[width=0.23 \textwidth]{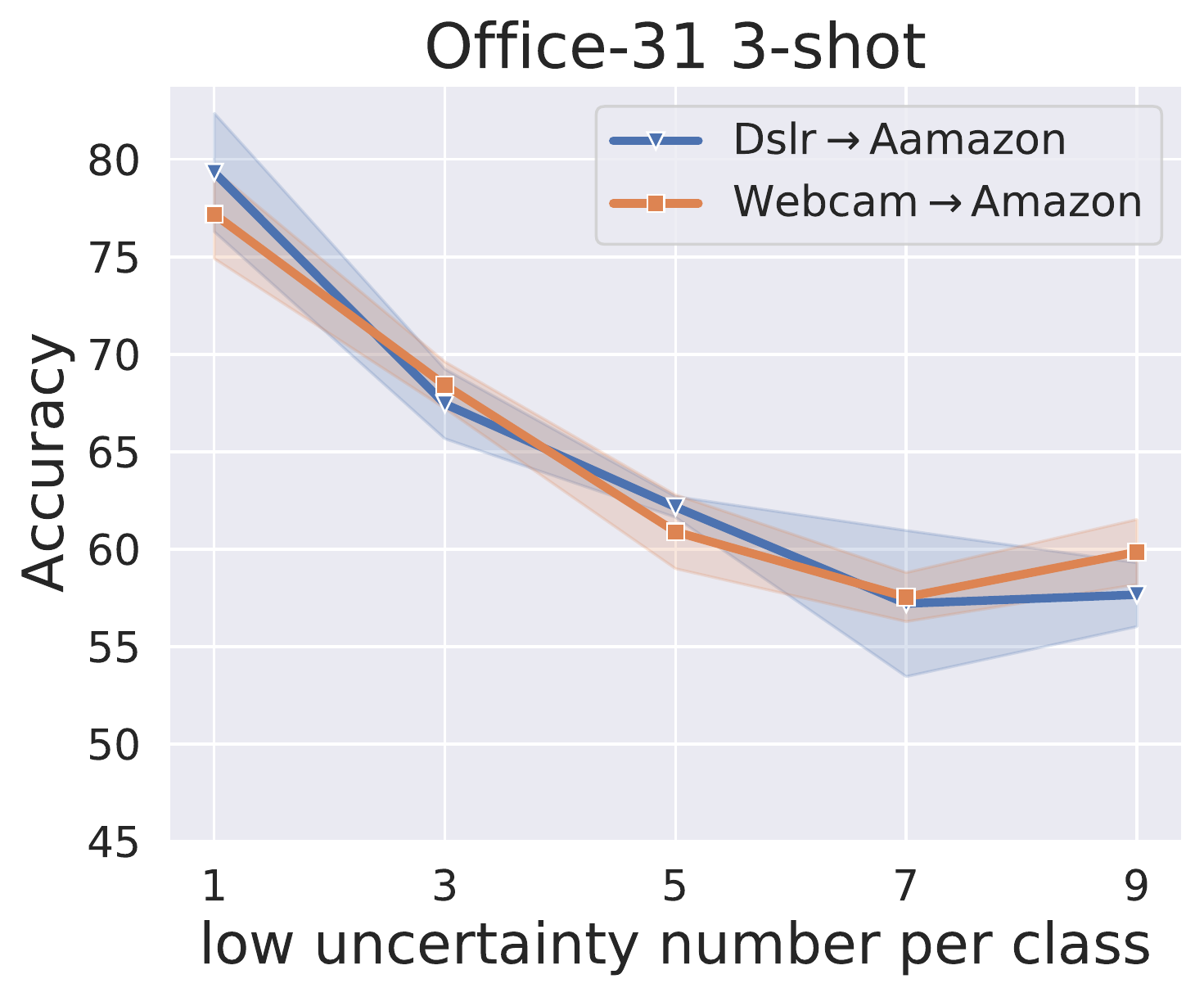}
}
\subfigure[\scriptsize{different $\lambda_{0}$ for 1-shot}]{
\label{fig:data5}
\centering
\includegraphics[width=0.23 \textwidth]{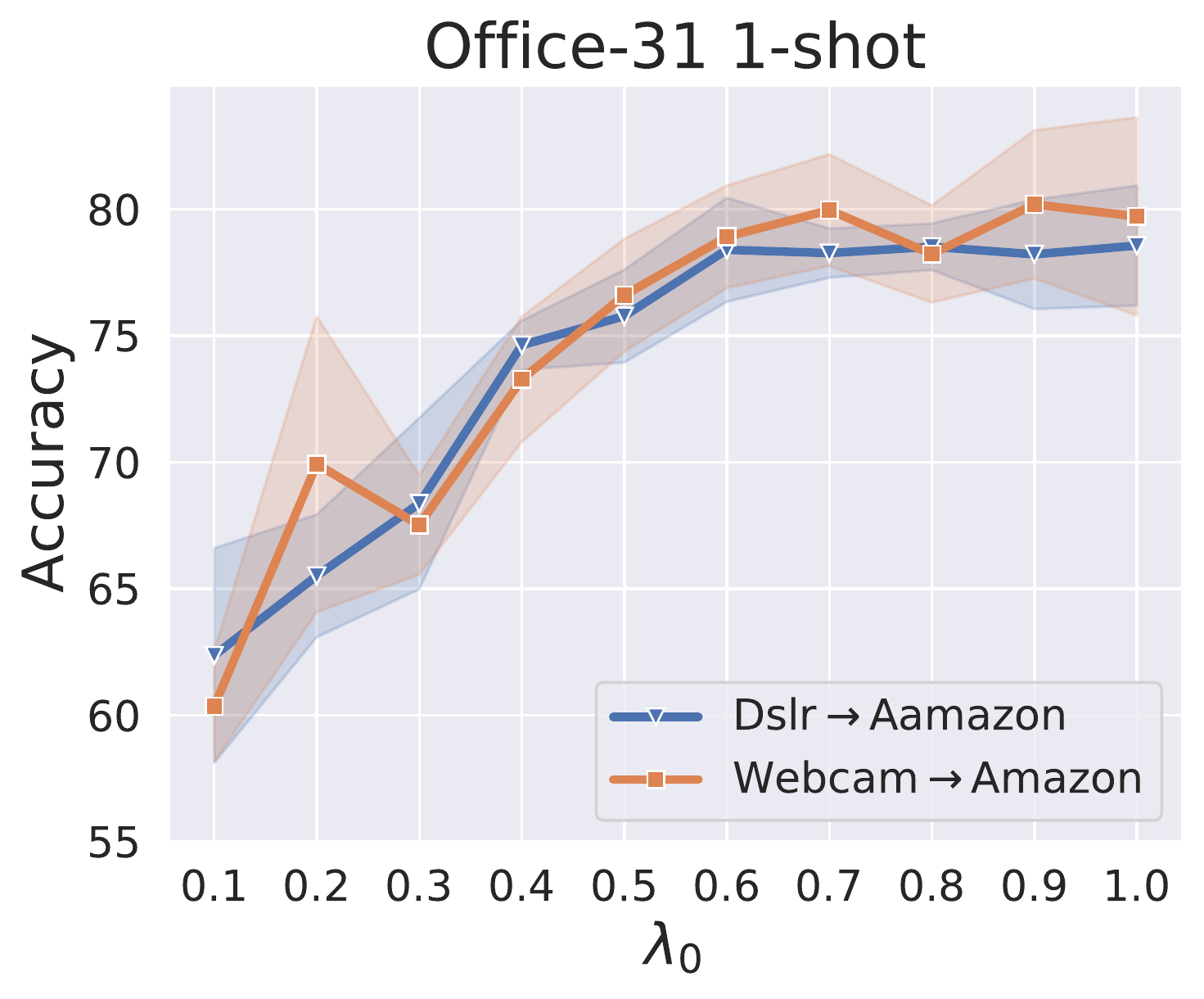}
}
\subfigure[\scriptsize{different $\hat{k}$ for 1-shot}]{
\label{fig:data6}
\centering
\includegraphics[width=0.23 \textwidth]{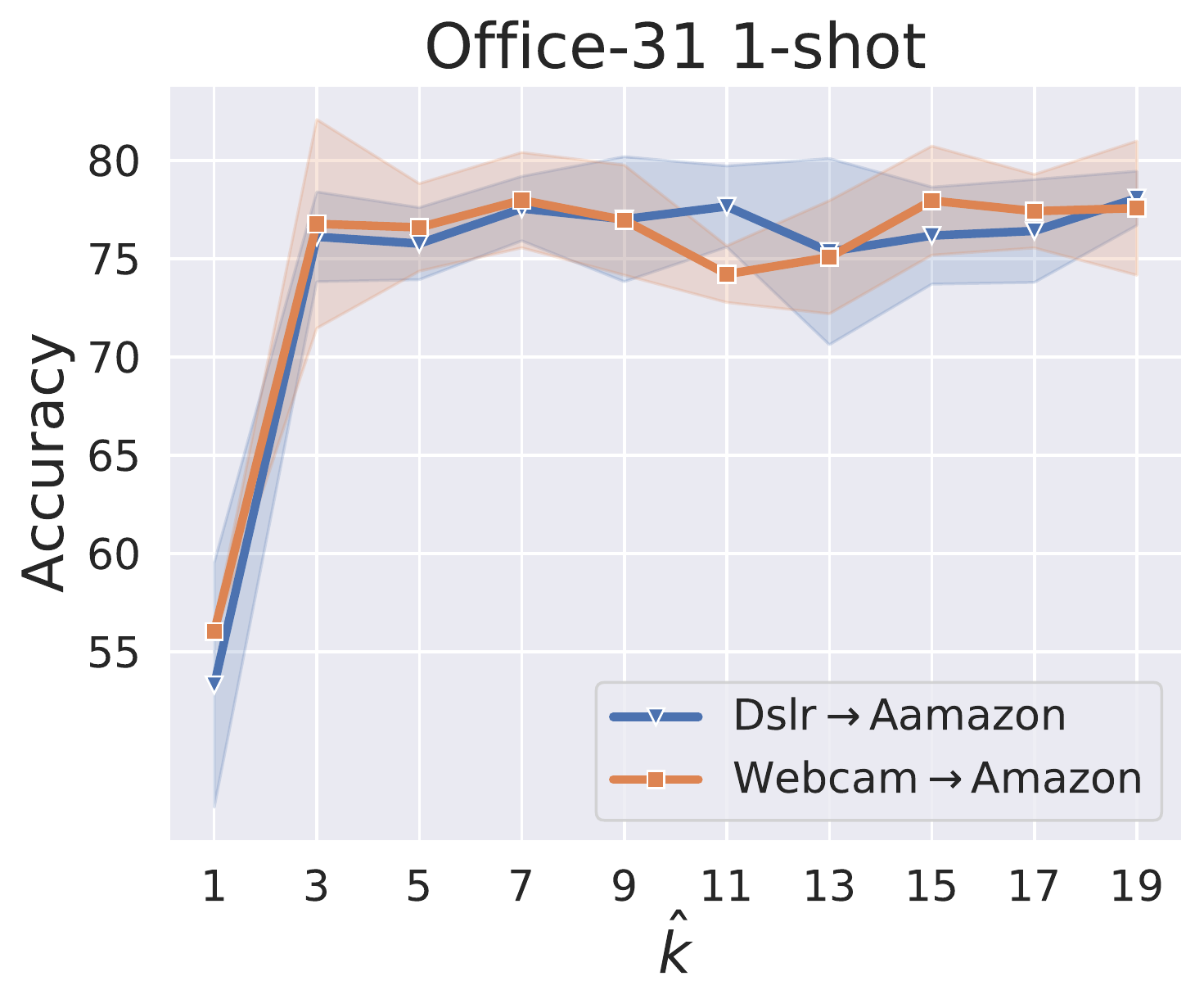}
}
\subfigure[\scriptsize{different shot on DomainNet}]{
\label{fig:data7}
\centering
\includegraphics[width=0.23 \textwidth]{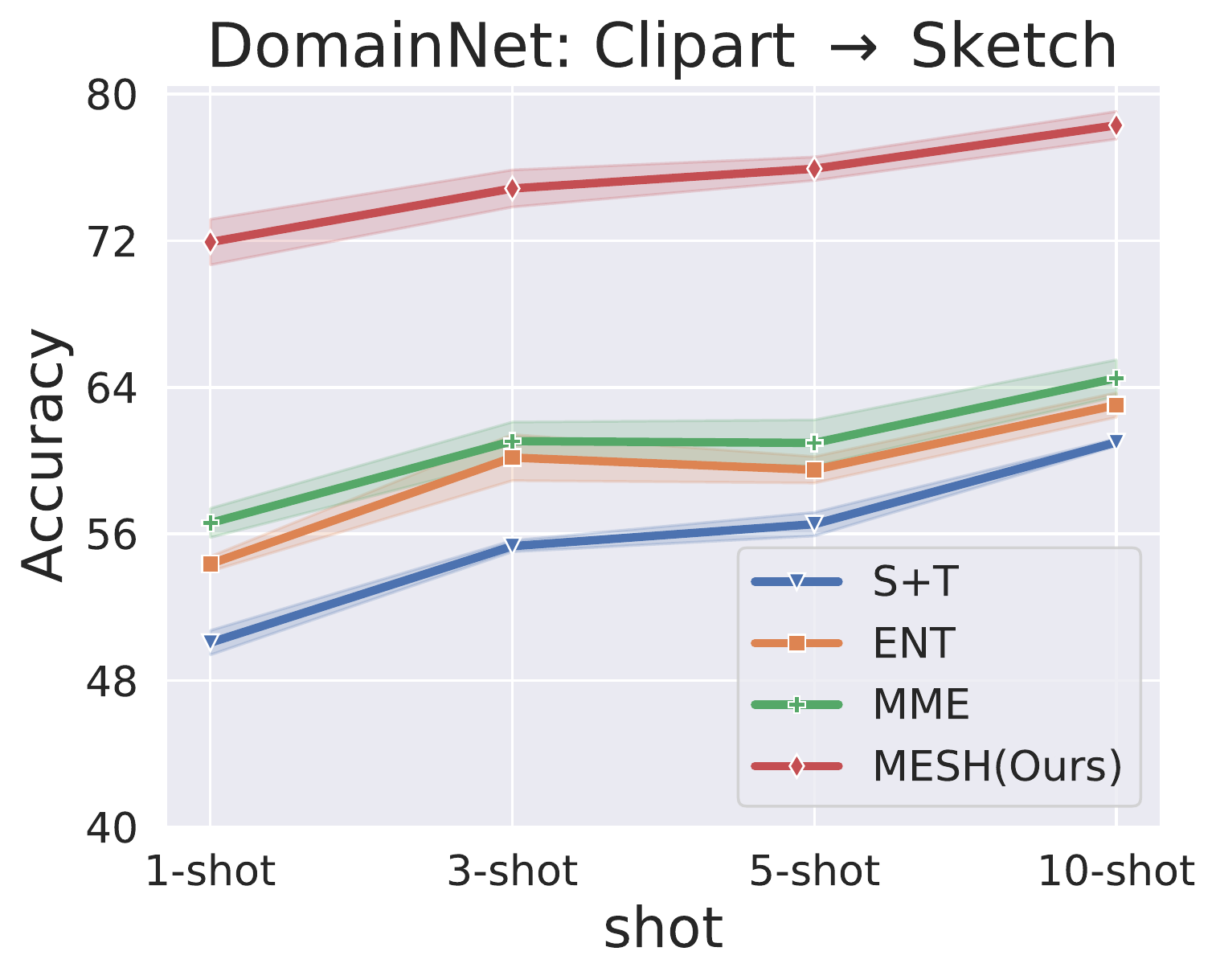}
}
\subfigure[\scriptsize{different shot on Office-Home}]{
\label{fig:data8}
\centering
\includegraphics[width=0.23 \textwidth]{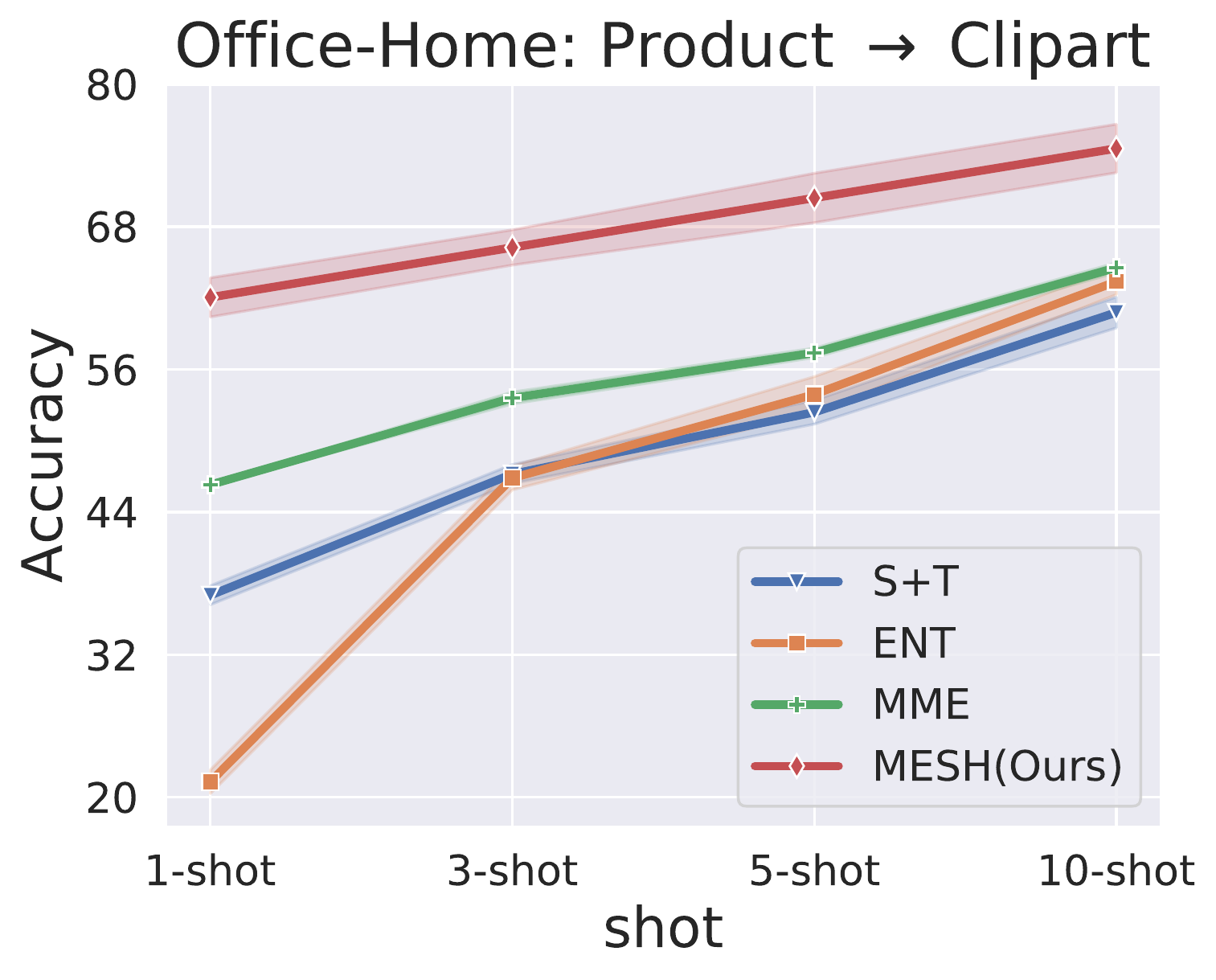}
}
\caption{Detail analyses on different hyper-parameters and backbones. The vertical axis is the test accuracies. n-shot denotes using n labeled target data for each class. We repeat three times and report the mean with a standard variance. (a)(b) Prediction on the selected low uncertainty data with or without \textbf{E}ntropy \textbf{M}inimization. (c)(d) Different low uncertainty examples for final performance. (e)Different weight parameters for training loss. (f) Different k-nearest neighbor number. (g)(h) Different shot for each class.}
\label{fig:multi_shot}
\end{figure*}

\textbf{Breaking the mutual enhancement training.} We build our framework without the low uncertainty augmentation (see \textbf{MESH-nA} in Table \ref{tab:office},\ref{tab:officehome},\ref{tab:domainnet}).  Once low uncertainty augmentation is removed, the enhancement is broken and the model suffers large performance loss. Furthermore, we want to find specific evidence proving that the entropy minimization can also boost our proposed label propagation method. We find that when removing the entropy minimization from our framework,  the quality of selected low uncertainty data suffers large degeneration (see Figure \ref{fig:data1}, \ref{fig:data2}). 
\begin{table}[!h]
\centering
\caption{Ablation experiment results on the Office-Home dataset. we evaluate the VGGNet based model with three labeled examples per target class. P: Product, C: Clipart. The first result (48.1\%) is reported by only using the labeled target data without any of the regularization terms. Note that the $\mathcal{L}_{ent}$ is label propagation loss without low uncertainty data.}
\begin{tabular}{ccccc|c} 
\toprule
& $\mathcal{L}_{ent} $ & $\mathcal{L}_{ps}$  & $\mathcal{L}_{vadv}$ & $\mathcal{L}_{div}$ &Accuracy \\
\midrule
\multirow{8}{*}{P-C} & & & & &48.1\\
\cline{2-6}
& \checkmark  & & & &48.8 \\
\cline{2-6}
 & &\checkmark & & &50.7 \\
\cline{2-6}
 & & &\checkmark & &48.9\\
\cline{2-6}
 & & & & \checkmark &43.9 \\
 \cline{2-6}
 & \checkmark & \checkmark& & &52.5 \\
\cline{2-6}
 & \checkmark & \checkmark& \checkmark & &53.7 \\
\cline{2-6}
 & \checkmark & \checkmark & \checkmark & \checkmark &55.8 \\
 
\bottomrule
\end{tabular}

\label{tab:ablation}
\end{table}

 \textbf{Effect of each regularization term.} We perform ablation experiments to provide more details of each regularization term. More specifically, we perform a domain adaptation task from Product to Clipart, where both the two domains belong to Office-Home. Our model gets 48.1\% classification accuracy only with 3 labeled target data using cross entropy loss $\mathcal{L}_{lab}$ (see Table \ref{tab:ablation}). 
Based on $\mathcal{L}_{lab}$,  each regularization term 
improves the adaptation performance, except $\mathcal{L}_{div}$. Because when only applying $\mathcal{L}_{div}$, the model is encouraged to make a balanced label assignment without further limitation for unlabeled target data, which breaks the discriminative representations of unlabeled data. Note that the $\mathcal{L}_{ps}$ is pseudo labeling loss without low uncertainty data.

\textbf{Hyper-parameters analyses.} The model's key hyper-parameters include the number of low uncertainty data, the weight parameter $\lambda_{0}$ and the nearest neighbor number $\hat{k}$. To determine the number of low uncertainty data, one strategy is to set a uncertainty threshold. However, the threshold is usually dependent on different datasets or classes of datasets. Instead,  we release the threshold and select the sample with lowest uncertainty for each target class. The reason behind it is propagating the pseudo labels with highest quality introduce least noises for the models. Figure \ref{fig:data3},  \ref{fig:data4} show that more low uncertainty examples leads to performance degradation. Then we conduct experiments using different $\lambda_{0}$.   Figure \ref{fig:data5} demonstrates the performance using different $\lambda_{0}$. When $\lambda_{0}$ is small, the label propagation can not improve the performance well.  We also investigate the effect brought by different $\hat{k}$. Figure \ref{fig:data6}  shows that when $\hat{k}=1$, the built nearest neighbor graph is too sparse to propagation label information. When $\hat{k}>1$, effective propagation can be performed. Finally, we perform multiple adaptation tasks on Office-Home and DomainNet with varied labeled target examples (see Figure \ref{fig:data7}, \ref{fig:data8}). Our method outperforms baselines in all scenarios when given different numbers of labeled target data. 

\section{Discussion and Conclusion}
In this paper, we proposed a source-free method for semi-supervised domain adaptation (SSDA). Compared to traditional SSDA methods, our method successfully protects the privacy of source domain data while ensuring efficiency and ease of implementation. 

Although our method gets satisfactory performance under the setting of hypothesis transfer,  there are still some challenging problems. For example, is hypothesis transfer always working in the SSDA problems?  Our insight is hypothesis transfer has better performance when given over-parameterized encoder networks and classifier networks. However, when given a lower capacity network for sufficient labeled target example, it is not sure that hypothesis transfer consistently outperforms the vanilla fine-tuning method. Besides that, is the mutual enhancement training still working for unsupervised domain adaptation?  How to further improve the training method without using hyperparameter $\lambda_{0}$? We leave these problems as future work.  

\bibliographystyle{ACM-Reference-Format}
\bibliography{sample-base}
\end{document}